\documentclass[letterpaper, 10 pt, conference]{ieeeconf}

\IEEEoverridecommandlockouts
\let\labelindent\relax
\usepackage[english]{babel}
\usepackage[protrusion=false,stretch=10,shrink=10]{microtype}
\usepackage{hyperref}
\usepackage{cite}
\usepackage{amsmath, amssymb, amsfonts}
\usepackage{graphicx}
\usepackage[font=footnotesize]{caption}
\usepackage{subcaption}
\usepackage{siunitx}
\usepackage{enumitem}
\usepackage{listings}
\usepackage{algorithmic}
\usepackage{booktabs}
\usepackage{textcomp}

\usepackage{xcolor}
\definecolor{mygreen}{rgb}{0.375, 0.411, 0.298}
\definecolor{myblue}{rgb}{0.278, 0.353, 0.445}

\begin{document}

\title{Optimization-Based Mechanical Perception for Peduncle Localization During Robotic Fruit Harvest}

\author{Miranda Cravetz$^1$, Cindy Grimm$^1$, Joseph R. Davidson$^1$
\thanks{This research is supported by the United States Department of Agriculture (USDA)-National Institute of Food and Agriculture (NIFA) through the Cyber Physical Systems program (Award No. 2020-67021-31525).}%
\thanks{$^1$Collaborative Robotics and Intelligent Systems (CoRIS) Institute, Oregon State University, Corvallis, OR 97331, USA. \newline
        {\tt\footnotesize \{cravetzm, grimmc, joseph.davidson\}@oregonstate.edu}}
}

\maketitle

\begin{abstract}

Rising global food demand and harsh working conditions make fruit harvest an important domain to automate. Peduncle localization is an important step for any automated fruit harvesting system, since fruit separation techniques are highly sensitive to peduncle location. Most work on peduncle localization has focused on computer vision, but peduncles can be difficult to visually access due to the cluttered nature of agricultural environments. Our work proposes an alternative method which relies on mechanical --- rather than visual --- perception to localize the peduncle. To estimate the location of this important plant feature, we fit wrench measurements from a wrist force/torque sensor to a physical model of the fruit-plant system, treating the fruit's attachment point as a parameter to be tuned. This method is performed inline as part of the fruit picking procedure. Using our orchard proxy for evaluation, we demonstrate that the technique is able to localize the peduncle within a median distance of \SI{3.8}{\cm} and median orientation error of \SI{16.8}{\degree}. 

\end{abstract}
\section{Introduction}
For the production of many specialty fruit crops, such as fresh market tree fruit, manual harvesting remains an unfortunate necessity. Harvesting is the most labor-intensive task of the annual production process, accounting for up to 48\% of all labor hours~\cite{Seavert2007EnterpriseRegion}. The work itself is unpleasant -- exactly the type of ``dirty, dangerous, and demanding'' work for which a robot is well-suited. Society currently faces a shortage of agricultural labor due, in part, to the nature of this work. There are also known health risks associated with harvesting work, such as the development of musculoskeletal disorders~\cite{Fathallah2010}. Furthermore, the amount of harvesting labor needed is expected to increase due to rising global food demand~\cite{van2021meta}. All of these concerns have led to increased interest in robotic harvesters both within the research community and the agriculture industry.

A major challenge in the realization of robotic fruit harvesters has been the development of adequate perception systems. Agricultural environments are filled with variation and occlusions which make perception difficult. A fruit may be obscured from view by leaves or an adjacent fruit, and may have a slightly different size and appearance from its neighbors. This has been a considerable challenge for researchers, who seek to use vision-based fruit and peduncle detection. Viewpoint selection has made advances in dealing with partial visual occlusions in agricultural environments \cite{Hemming_2014 , vanEssen_2022}, but is still limited by the modality of vision itself. For important plant features that are completely visually obscured, a different approach is needed. Recent work on mechanical perception for peduncle localization holds promise, but has only been applied to tomato plants and requires an isolated exploratory procedure which adds time to the harvest process~\cite{Senden_2022}. 

\begin{figure}[t]
    \centering
    \includegraphics[width =0.9\linewidth]{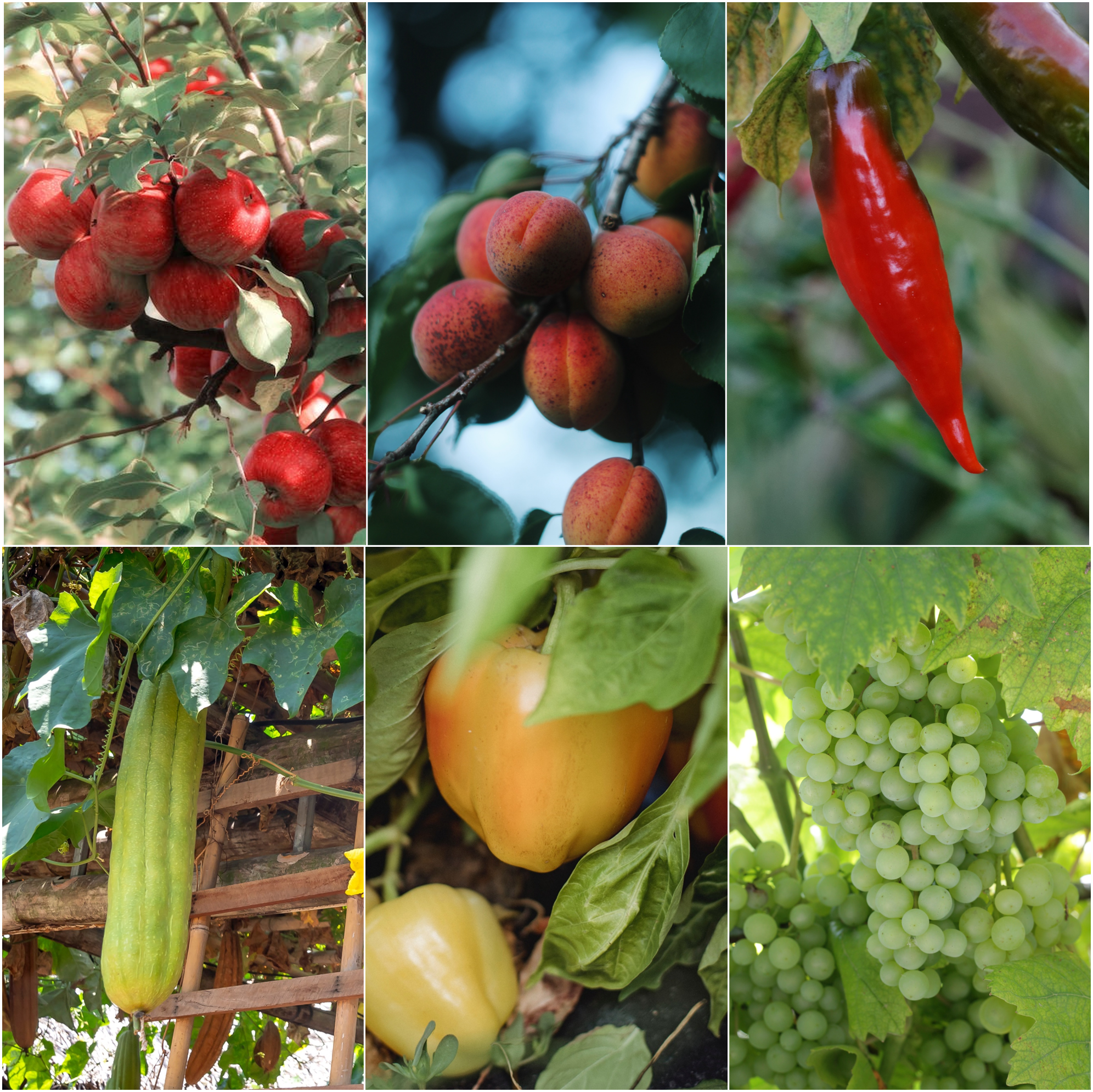}
    \caption{Several examples of fruit with visually-occluded peduncles. The peduncle is an important fruit feature -- optimal picking patterns often depend on knowledge of the peduncle's position.}
    \label{fig:abstract_graphic}
    \vspace{-5mm}
\end{figure}

In this paper, we propose a during-pick procedure for peduncle attachment point localization. The peduncle is the fruit-bearing or fruit-cluster-bearing stem of a plant, and its attachment point to the plant is a feature that is both frequently fully visually occluded (Figure \ref{fig:abstract_graphic}) and critically important to know in order to cleanly separate the fruit from the plant. We observe that when the robot displaces a fruit, the plant's foliage resists this motion with a restoration force. Since this restoration force is dependent on the plant's architecture, it is intuitive that information about the plant can be captured by analyzing the force. To localize the peduncle attachment point, we fit the data from a Force/Torque (F/T) sensor in a robot's wrist (a common sensor configuration for modern robots) to a physical model of the fruit-peduncle-plant system, treating the peduncle attachment point as a model parameter to be tuned. Similar physical modeling approaches exist in the literature for defining object articulations and environmental contact conditions~\cite{subramani2018recognizing, cai2020inferring}. However our method deviates from this approach in that we use a quasi-static rather than purely kinematic model and in that the robot exchanges significant energy with the environment during interaction.

Our proposed peduncle estimation can be executed in the time between grasping the fruit and executing the picking motion (also called the picking pattern), adding minimal time to the harvesting process. We validate our procedure using stem-loading data collected from our proxy-orchard environment presented in~\cite{velasquez2022predicting} using our robotic harvesting system presented in \cite{Dischinger2021Towards}. Our results show that the attachment point location can be estimated to within a median of 3.8 cm of the measured location, or a median peduncle angle difference of \SI{16.8}{\degree}. In Section \ref{sec:relwork} we discuss previous research related to our contribution. Section \ref{sec:methods} describes in detail our proposed procedure and our validation methodology and Section \ref{sec:results} documents our results. Finally Section \ref{sec:disc} discusses the implications of our results and our contribution to the overall literature on robotic harvesting.

\section{Related Work}
\label{sec:relwork}
Robotic harvesters have been the subject of research interest since the 1980's. As such, there is an extensive body of work in this area, with one 2015 review citing more than 70 robotic harvesting systems, 64 of which were developed for fruit~\cite{Bachche_2015}. Another review \cite{bac2014harvesting} found that, despite these research efforts, the performance of harvesting robots has not improved since their inception. The reason cited for this lack of improvement is the complexity of the agricultural environment. Variation within the plants and fruits, the climate, and the growing conditions all add to this complexity and pose a major challenge for robotic perception.

\begin{figure}[h]
    \centering
    \includegraphics[width = \linewidth]{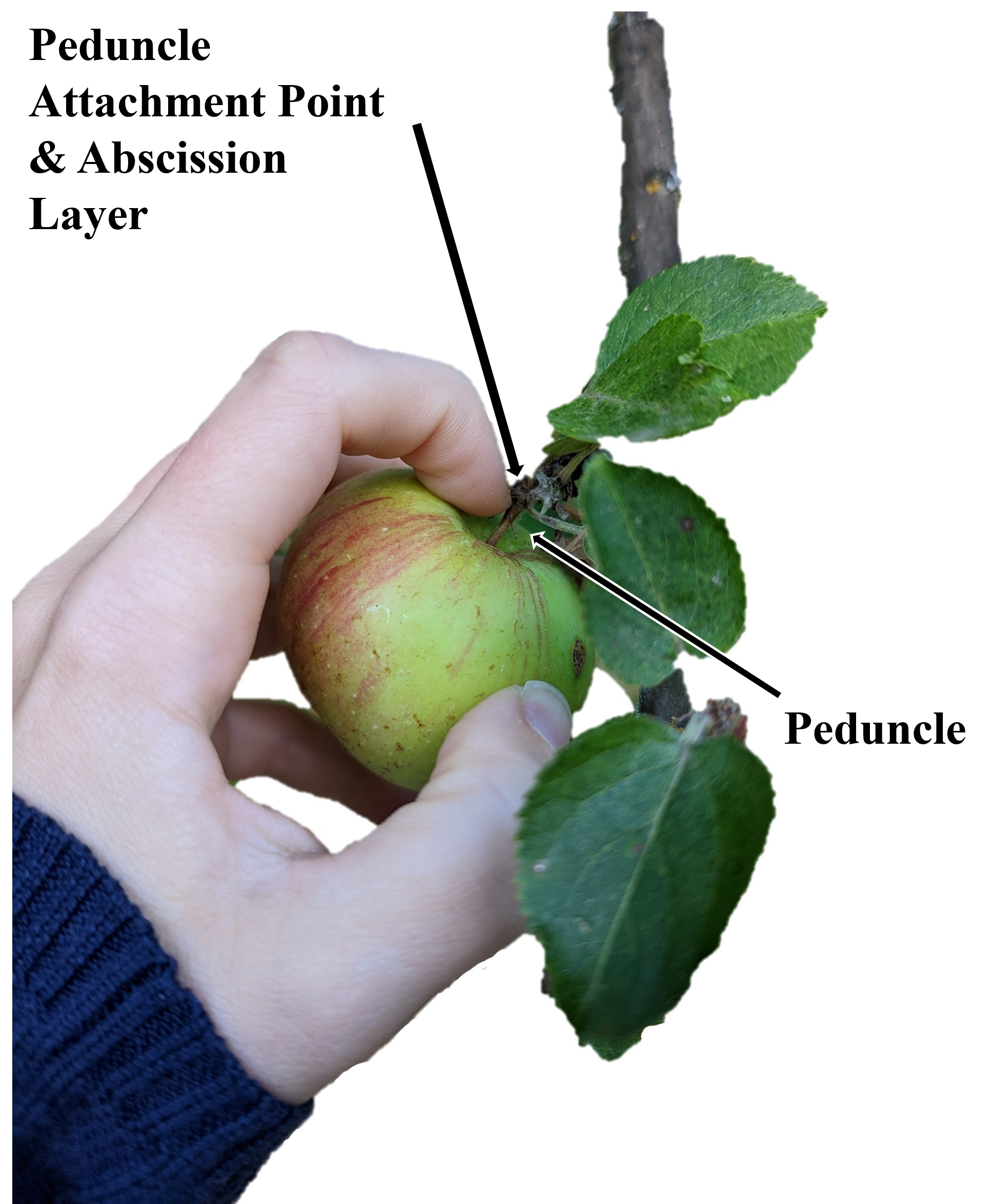}
    \caption{A human hand grasping an apple, with relevant plant features labelled. Human apple pickers separate the fruit by applying a bending force directly to the stem. This is the ideal way to create shear stress at the abscission layer, which for apples is located at the attachment point between the peduncle and the fruiting spur. When force cannot be applied to the stem directly, the ideal picking strategy becomes gently tensioning the stem and rocking the apple in an arc about the abscission layer.}
    \label{fig:plant_anatomy}
    \vspace{-5mm}
\end{figure}

\begin{figure*}[!t]
    \centering
    \includegraphics[width=0.9\linewidth]{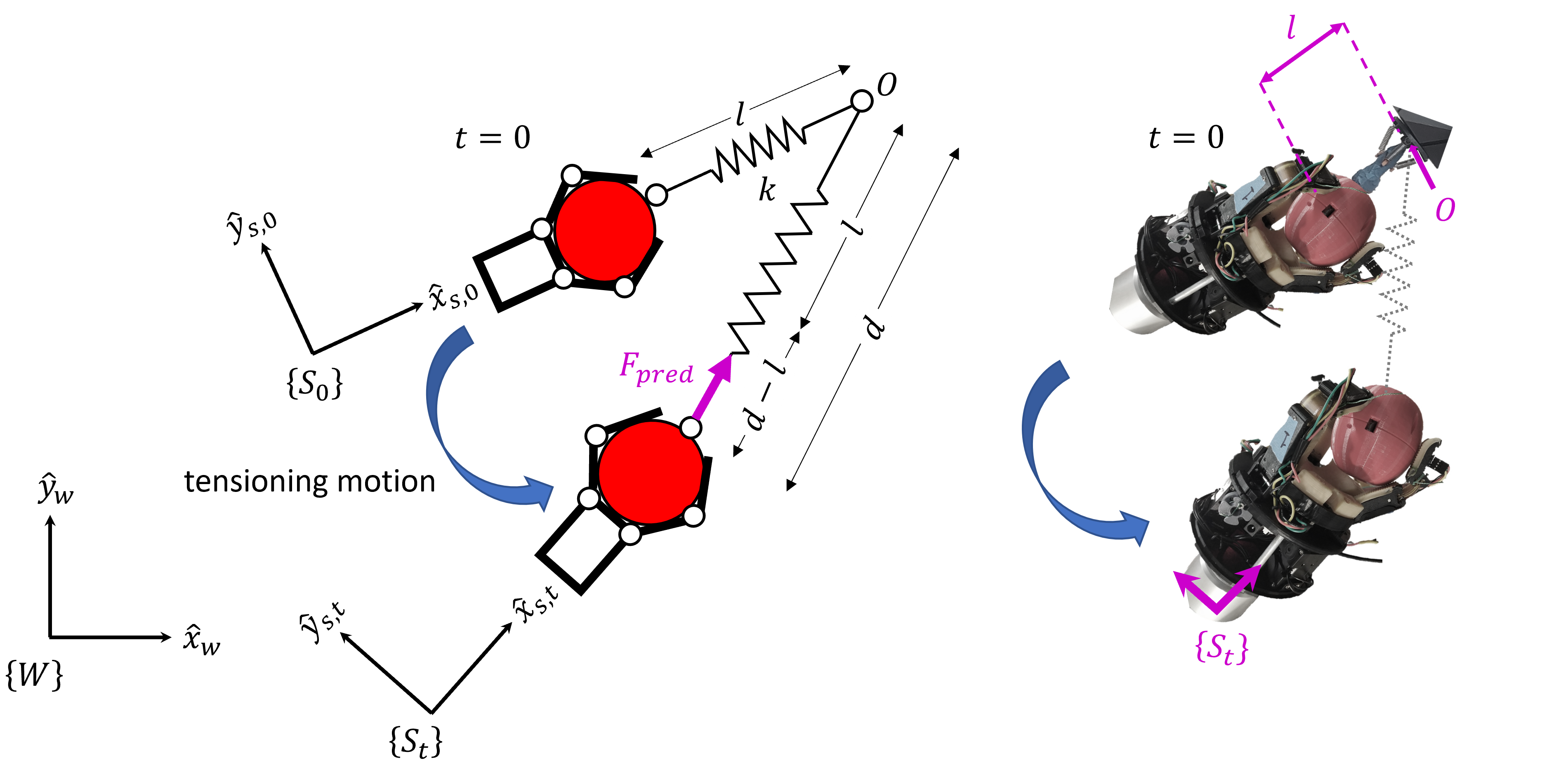}
    \caption{The physical model used in this paper. Left: As the robot hand holding the fruit moves from its initial position (top) at \(t=0\) to a new position (bottom) at some time \(t\), the wrist F/T sensor records a change in tension. We model the peduncle as a spring with length $l$, spring constant $k$. The force is recorded in the sensor frame, \(\left \{ S \right \}\) , but can be transformed into the world frame, \(\left \{ W \right \}\). Right: we show the physical robot hand, grasping the fruit, moving in a similar motion.}
    \label{fig:model}
    \vspace{-5mm}

\end{figure*}

To separate fruits by pulling, the picking motion must transfer force to the abscission layer of the fruit without exerting excess force on any other part of the plant (Figure \ref{fig:plant_anatomy}). This requires information about the peduncle location and orientation. Both hand picking studies and advanced models have shown that, in apples, picking motions which are not sensitive to peduncle attachment location are prone to fruit damage from both bruising and stem separation \cite{li2016characterizing,bu2020experimental}. Similar results have been shown for tomatoes, where the excessive force caused by improperly loading the pedicel and peduncle can lead to peduncle breakage and thereby loss of entire fruit clusters \cite{liu2020experimental}.

Peduncle localization is also important for separating fruits via cutting through the peduncle. In this case, incorrect location estimation can cause damage to the plant or fruit. Peduncle localization for this purpose has been studied in tomatoes \cite{Senden_2022} and in sweet peppers \cite{lehnert2020performance}. However, these localization strategies may not be feasible for commercial implementation. In~\cite{Senden_2022} an exploratory procedure is used that requires exciting the tomato plant at frequencies as low as \SI{0.5}{\hertz}, resulting in a procedure execution time of 40 minutes. Although the authors make suggestions for how to improve the speed, it remains unclear how much the execution time can be reduced. In~\cite{lehnert2020performance}, the researchers use a vision based localization system which requires modifications to the plant and greenhouse environment, such as leaf removal, to be effective. These modifications take manual labor to perform, which may not be practical in commercial settings.

For our approach to peduncle localization, we treat the connection between the parts of the plant as an environmental constraint acting on the grasped fruit. Such environmental constraints have been addressed in many areas of manipulation literature, including articulated object manipulation, contact estimation, and physical human interaction \cite{subramani2018recognizing, cai2020inferring,jain2020learning,ansari2020human}. Parameter estimation for a physical model is often used in these areas to deconstruct the manipulation task into an allowable motion direction and a force direction. Our work extends these physical modeling techniques to an environmental interaction where the force is in the same direction as the motion (i.e. energy is exchanged) and introduces them to the agricultural setting.

\section{Methods}
\label{sec:methods}

\subsection{Physical Model}


We begin by describing the physical model we use for our localization procedure. \textit{The key idea is that since the force predicted by the model depends on the location of the fruit's peduncle, we can use force measurements to determine this location}. By tuning the peduncle attachment location until the modeled force is accurate, we can retrieve the peduncle location using just the force data from the robot's wrist.

We model (Figure~\ref{fig:model}) the fruit's peduncle as an idealized linear spring with stiffness $k$ in translation and infinite compliance in bending (i.e. zero rotational stiffness). The peduncle couples the fruit to the tree -- we represent the tree-side attachment point as a frictionless spherical joint, and the apple-side attachment point as rigid. We define a point \(r_O = [x_O, y_O, z_O] ^T \) in the world frame \(\left \{ W \right \}\) at the peduncle's attachment to the tree, \textit{which is considered fixed in space} (shown as O in Figure~\ref{fig:model}). We also define a time dependent point \(r_{a,t} = [x_{a,t}, y_{a,t}, z_{a,t}] ^T \) which represents the location of the apple at time \(t\). 

\begin{figure*}[!t]
    \centering    \includegraphics[width = \linewidth]{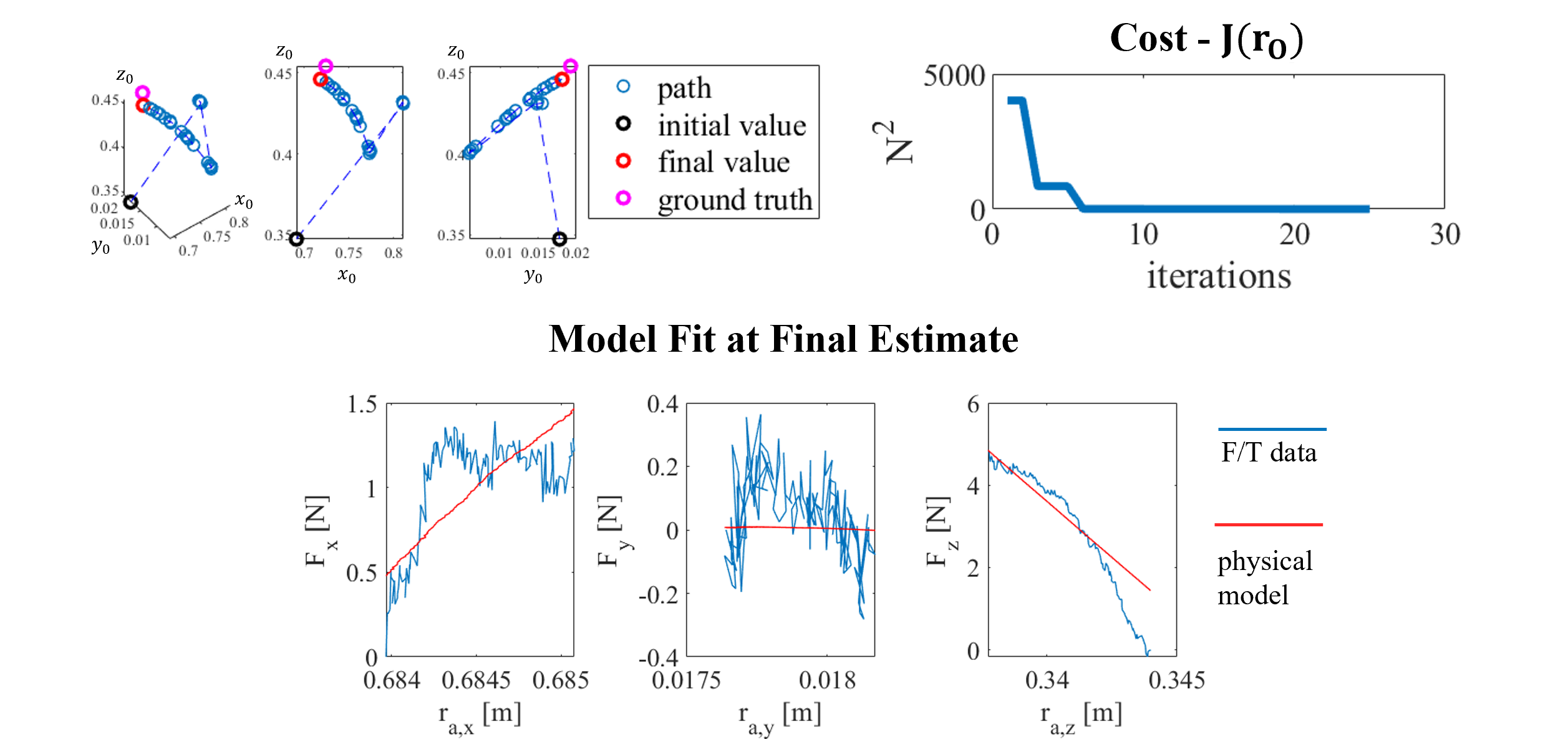}
    \caption{An example of the optimization process for a single case. The progression of the optimization over several iterations is shown on top, with the path taken by the optimization algorithm on the left and the cost at each iteration on the right. At the bottom is the final fit achieved by the optimization process.}
    \label{fig:optimization_example}
    \vspace{-5mm}
\end{figure*}

\begin{figure}[h]
    \centering    
    \includegraphics[width = \linewidth]{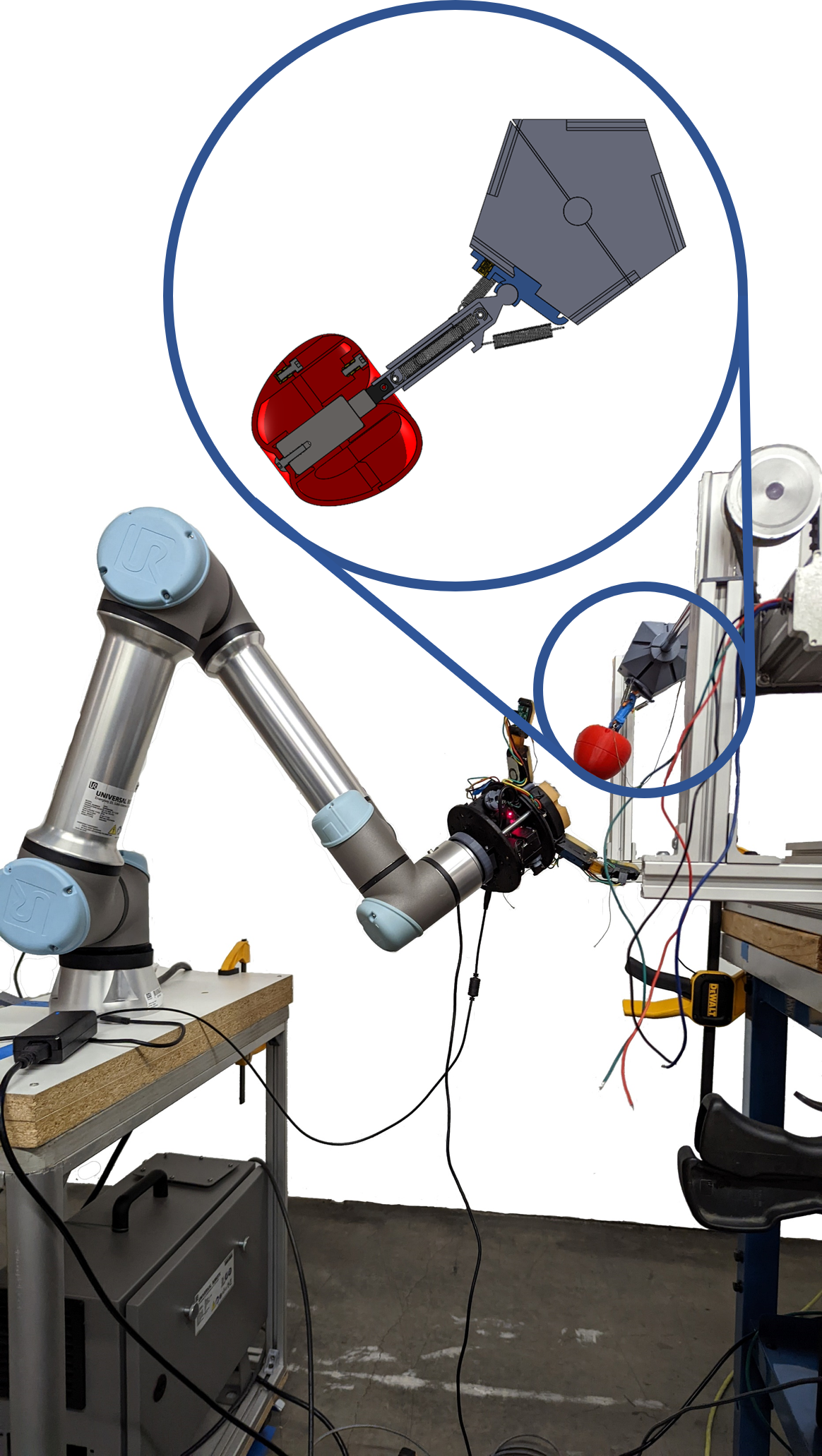}
    \caption{The data collection setup used in this study. The robot is shown on the left and the proxy orchard environment on the right. Inset is a render of a cross section view of the proxy orchard's spring mechanism.}
    \label{fig:setup}
    \vspace{-5mm}
\end{figure}

When the robot completes its initial grasp of the fruit (i.e. \textit{t} = 0) the system is in static equilibrium with the spring at its resting length \(l\), which we treat as known. This can also be expressed as \(\left \| r_O - r_{a,0} \right \| = l\). As the robot loads the stem by moving the fruit away from this equilibrium, the spring elongates to a new length (this elongation captures displacement of plant vegetation during real apple picks). This spring displacement generates a force \(F_{pred} \in \mathbb{R}^3\). This force can be found by \[F_{pred} = k(\left \| d \right \|-l))\frac{d}{\left \| d \right \|}\] where \(d = r_O-r_{a,t}\) is the vector from the apple to the peduncle attachment point.

To determine the location of \(r_a,t \) we define a frame \(\left \{ S \right \}\) attached to the F/T sensor in the robot's wrist. The transformation between this frame and \(\left \{ W \right \}\) can be extracted from the robot's forward kinematics. \textit{We consider \(r_a \) as fixed in this sensor frame throughout time (i.e. the fruit is held in a perfectly rigid grasp)}.

\begin{figure}[h]
    \centering
    \includegraphics[width = 0.9\linewidth]{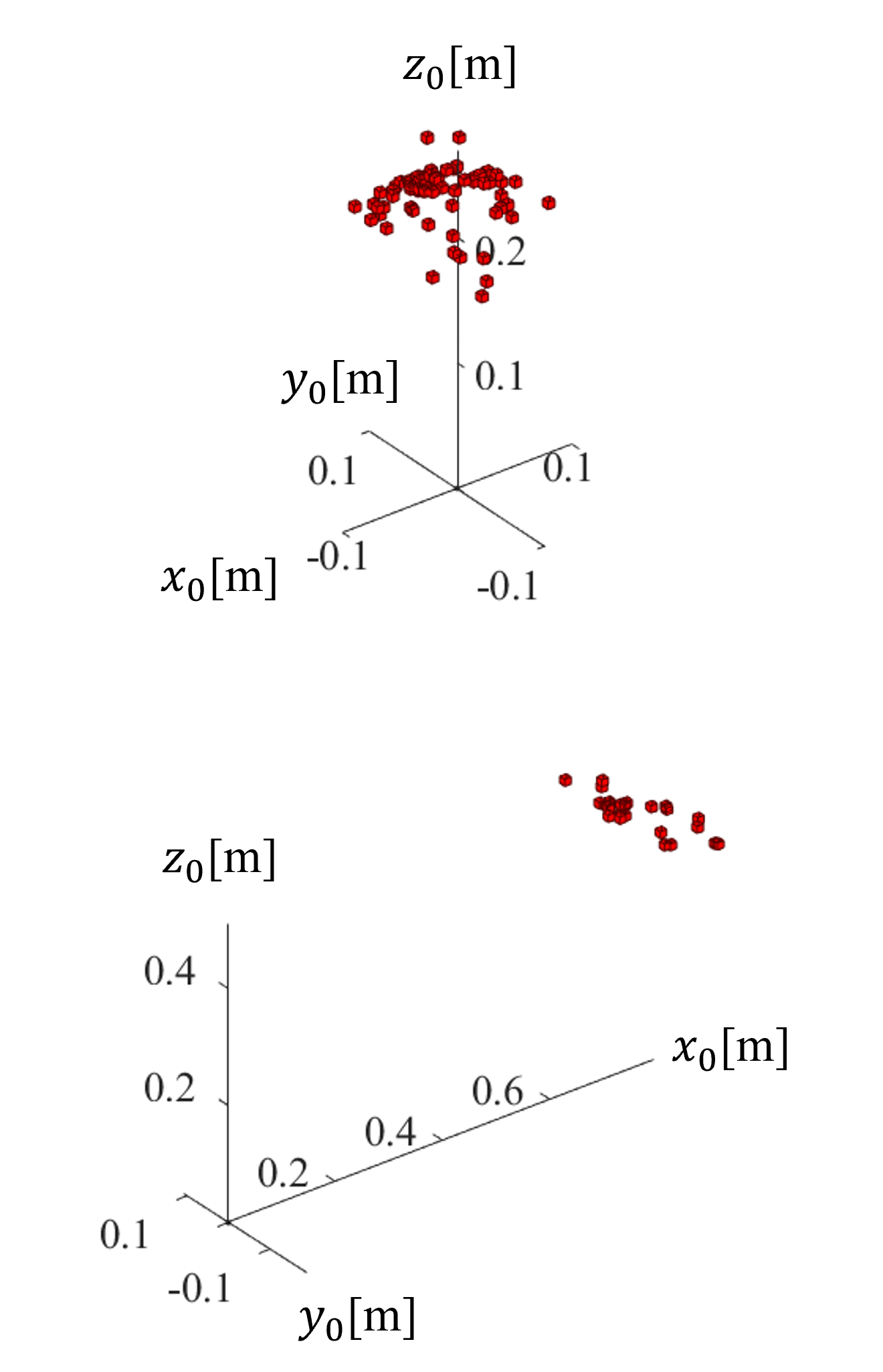}
    \caption{The positions of the proxy peduncle attachment point shown in frames \(\left \{ S_{0} \right \}\) (top) and \(\left \{ W \right \}\) (bottom)} 
    \label{fig:joint_locations}
    \vspace{-3mm}
\end{figure}

\begin{figure*}[h]
    \centering    \includegraphics[width = \linewidth]{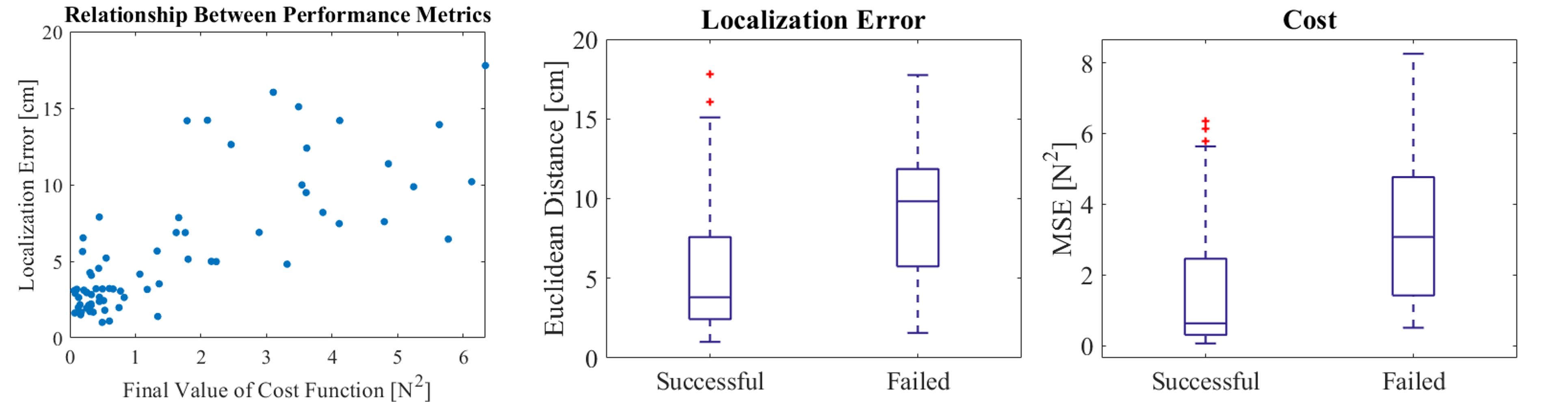}
    \caption{Overall performance of the optimization procedure. For detailed values, refer to Table \ref{tab:metrics}.}
    \label{fig:performance_metrics}
    \vspace{-5mm}
\end{figure*}

\subsection{Optimization Procedure}

To tune the peduncle location within the model, we use well-established optimization techniques. Optimization iteratively adjusts the location in a way that reduces the disparity between the predicted force from the model and the measured force from the F/T sensor (Figure \ref{fig:optimization_example}). We now describe the specific optimization procedure. 

At each time step, we calculate the error between the force generated by our model, \(F_{pred}\) and the force measured at the F/T sensor transformed into \(\left \{ W \right \}\)  using an adjoint mapping, \(F_{meas}\). We then minimize the Mean Squared Error (MSE) over the timeseries subject to the constraint that the stem is in tension at every timestep. That is, we define the constrained optimization problem as:


\begin{align*}
min(\{J(r_O)=\frac{1}{n}\sum_{t=0}^{n}\left \| F_{meas,t}-F_{pred,t} \right \|^2 \}\\ \;  | \;  l - \left \| d_t \right \| \leq 0 \text { for all t} )
\end{align*}

The optimization was performed using Sequential Quadratic Programming in MATLAB's Optimization Toolbox. The initial estimate for \(r_O\) was set as the initial location of the fruit, \(r_{a,0}\), plus a small offset to avoid cost function evaluation failure. Gradients for both the cost function and the constraint functions were provided to the optimization solver. The convergence criteria was tightened until the MSE was within \SI{5}{\newton\squared} - up to a maximum of five times. The tightening was achieved by reseeding the optimization algorithm with the output of the previous iteration as the initial conditions while keeping the \textit{relative} convergence criteria the same. The final value of \(r_O\) at the end of the optimization process is used as the prediction of the peduncle attachment point location.

\subsection{Data collection and processing}

Data for evaluating our technique was collected using our custom apple gripper, a UR5e industrial robot arm (Universal Robots, Odense, Denmark), and our proxy orchard environment (Figure \ref{fig:setup}). This proxy environment allows us to collect sufficiently large datasets, even outside of the harvest season. It mimics the mechanics of a fruit tree using a spring system, the primary spring of which has a manufacturer listed stiffness of \SI{632}{\newton\per\meter}, which we used as the spring stiffness for our model. First, the attachment location was measured using a probe attached to the robot. The range of measured attachment point locations sampled can be seen in Figure \ref{fig:joint_locations}. Then, the robot chose and assumed a random end-effector orientation. These orientations were constrained to a quarter sphere, where the elevation ranged from horizontal to opposite gravity and azimuth angle ranged from the positive to the negative y-axis of the robot's base frame.  

A researcher then manually drove the robot such that the apple was within reach of the gripper, maintaining the random orientation as closely as possible. After that the gripper closed and the robot pulled back 15 cm along the normal axis of the palm. The researcher labeled each trial as successful if the artificial fruit was separated and as failed if the fruit escaped the hand. Picking trials in which the grasp failed were disregarded and the remaining 70 trials were analyzed. For each picking trial, the data from when the robot started to pull back until a 5N tensioning force was achieved were used to fit the physical model. This data sample represents what would be available in the field after a small motion to tension the stem and before executing a pick pattern. The 5N load was achieved in an average time of 0.45 seconds. The sensors on the robot were sampled at \SI{500}{\hertz}, so the average number of samples passed to the optimization software was 226. No negative relationship was observed between speed of the tensioning motion and quality of the optimization performance, which indicates that the robot's speed could be increased to reduce the time taken by this procedure.
\section{Results}
\label{sec:results}



\begin{figure}
    \centering
    \includegraphics[width = 0.9\linewidth]{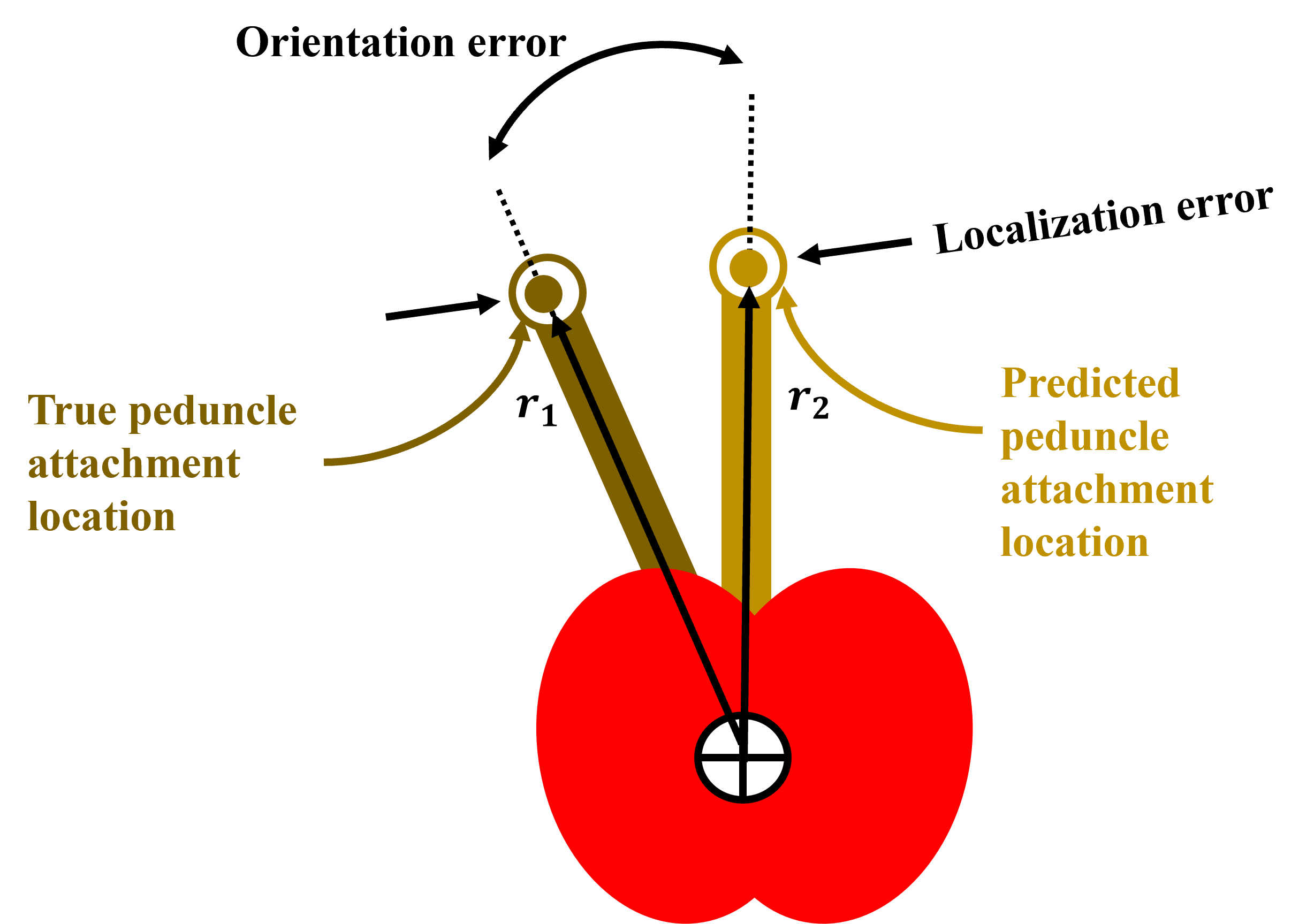}
    \caption{A sketch demonstrating how we calculate error in the peduncle location and orientation. Vectors $r_1$ and $r_2$ point from the fruit's center to the true and predicted peduncle attachment point locations, respectively. Their difference is the error in location, and the angle between them is the error in orientation.}
    \label{fig:metrics_sketch}
    \vspace{-5mm}
\end{figure}

For each of the 70 picks, we ran our optimization to predict \(r_O\) and recorded the metrics shown in Table \ref{tab:metrics} and defined here:

\begin{itemize}
    \item Final MSE: This is the value of the cost function (\(J(r_O)\)) evaluated at the final estimate for \(r_O\) as determined by the optimization procedure. A smaller final MSE means that the optimization was able to achieve a better fit between the model and the data.
    \item Localization error: This is the Euclidean distance between the predicted value for \(r_O\) as determined by the optimization procedure and the measured value of \(r_O\) using a probe attached to the robot (Section \ref{sec:methods}). 
    \item Orientation error: This is the angle between two vectors both pointing from the initial location of the fruit, \(r_{a,0}\), to either the predicted or measured location of \(r_O\). We calculate this angle as \(\arctan (\frac{\left \| r_1 \times r_2 \right \|}{r_1\cdot r_2})\) where \(r_1\) and \(r_2\) are the two vectors (Figure \ref{fig:metrics_sketch}).
    \item Optimization runtime: This is the time taken to run the optimization algorithm, including all restarts with tighter convergence criteria. The procedure was performed via MATLAB on a device with an Intel i7 processor.
\end{itemize}

The optimization was able to find a close solution for most picking trials. The median MSE between the data and the model was  \SI{0.63}{\newton\squared}. For cases where the MSE was under the median, the mean localization error was \SI{3.0}{cm}, compared to an overall mean of \SI{5.6}{cm}. Similarly, when the model proved to be a worse approximation of the data, the localization error increased (Figure \ref{fig:performance_metrics}). 

Excluding outliers, the average runtime was \SI{0.06}{\second}. Combining this with the average tensioning time of \SI{0.45}{\second}, the total total average time taken to perform the localization is \SI{0.51}{\second}. There were three cases considered outliers. These were cases where the optimization either failed to find a solution or performed many iterations without improving the fit. The impact of such cases could be reduced by limiting the maximum number of iterations taken by the optimization algorithm. This study allowed for a maximum of 10,000 iterations, but the maximum number of iterations taken outside of the outlier cases was only 169. A limit on iterations that is closer to this value would prevent failure cases from consuming impractical amounts of time. 

\begin{table}[]
\caption{Detailed performance metrics}
\begin{tabular}{|l|l|l|l|l|}
\hline
Metric                      & Median & IQR  & Mean & STD  \\ \hline
Final MSE {[}\SI{}{\newton\squared}{]}            & 0.63   & 2.15 & 1.6  & 1.76 \\ \hline
Localization Error {[}\SI{}{\cm}{]}   & 3.8    & 5.2  & 5.6  & 4.3  \\ \hline
Orientation Error [degrees] & 16.8 & 28.3 & 26.8 & 24.0 \\ 
\hline
Optimization Runtime {[}\SI{}{\second}{]} & 0.06   & 0.03 & 0.06 & 0.03 \\ \hline
\end{tabular}
\label{tab:metrics}
\end{table}

\begin{figure}[h]
    \centering    
    \includegraphics[width = \linewidth]{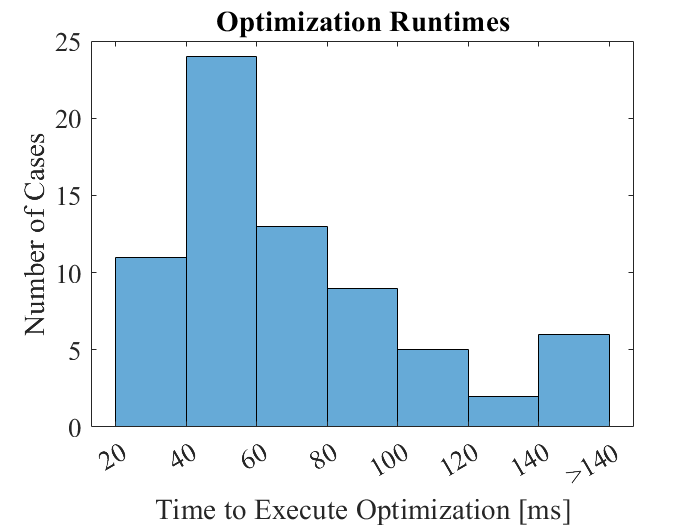}
    \caption{Runtimes of the optimization algorithm.}
    \label{fig:runtimes}
    \vspace{-5mm}
\end{figure}
\section{Discussion}
\label{sec:disc}

Our results imply that our proposed localization method is most accurate when the model closely describes the data. This is seen by the relationship in Figure \ref{fig:performance_metrics} between the error in the model and the error in the predicted position. Our model relies on several assumptions, namely:

\begin{itemize}
    \item The robot's grasp is a perfectly rigid power grasp, with the fruit in contact with the robot's palm.
    \item The attachment point between peduncle and branch remains perfectly fixed in space throughout the picking process.
    \item No energy is lost during the interaction due to friction or other sources.
\end{itemize}

The violation of these assumptions may be responsible for the ill fit between model and data seen in some of the trials. Our passively compliant hand often flexes its fingers during loading from a pick, which violates our first assumption. Compliance and friction within the proxy orchard system may also lead to violations of the other two assumptions. To test how much violation of our assumptions affects the optimized fit of the model, we looked to cases where the grasp eventually failed. These cases represent a large violation of our first assumption, where the fingers flexed enough that the proxy fruit was lost from the grasp. In 35 of the picking trials labeled as failure, a 5N tensioning force was achieved before the grasp failed. If we perform our optimization procedure on these cases, we find the performance becomes significantly worse. The median localization error increases to \SI{9.8}{cm} from \SI{3.8}{cm} for the successful cases (Figure \ref{fig:performance_metrics}). Similarly, the median final MSE between the model and data becomes \SI{3.07}{\newton\squared} compared to the previousy observed \SI{0.63}{\newton\squared}. Two-sample t-tests across classes on the two metrics reject the null hypothesis with \(p=6.4 \times 10^{-5}\) for localization error and \(p=5.2 \times 10^{-5}\) for final optimization cost. This can be taken as evidence of a link between violations of the model assumptions and poor performance outcomes.

Reducing the error in the model can be achieved in two ways. Either the data collection can be modified to more closely fit the assumptions, or the model can be modified to capture more of the mechanics. The former can be achieved by using a gripper which is able to achieve a more consistently rigid grasp, which would make the assumption of a rigid grasp more valid. Similarly, the model could be modified to include a variable estimate of the location of the fruit relative to the sensor, rather than considering the hand and fruit as a rigid system. 

Our model works well for our proxy orchard environment, which has been shown to have comparable mechanics to an apple tree \cite{velasquez2022predicting}. In the field, applying our model would require estimating system properties which are known to us for the proxy orchard system, such as the peduncle stiffness. This stiffness could either be measured directly by uniaxial pull tests or indirectly by estimating the peduncle width. Additionally, while the model likely works well for other tree fruits, modifications to the model may be needed to extend this work to more dissimilar crops. Tree branches are relatively stiff, making bulk compliance a good approximation of peduncle compliance. In many other cases, it may be necessary to consider the bending stress within the plant rather than just the bulk compliance, as done in~\cite{liu2020experimental}. 
\section{Conclusion}
\label{sec:concl}

This work demonstrates the feasibility of using mechanical perception to localize a fruit's peduncle during the picking process. This method holds promise for localizing this important plant feature in cases where computer vision is insufficient to do so, while contributing minimally to the time needed to harvest the fruit. This estimate can potentially be fed into a controller to perform advanced manipulation of the fruit, something that has been an unrealized goal of the agricultural robotics community. Future modifications which improve the agreement between the physical model and the real world will likely increase the accuracy achieved by this new technique, making it more suitable for this purpose.
\section*{Acknowledgments}

We thank Alejandro Velasquez for the use of his motion planning and data logging software.

\bibliographystyle{IEEEtran}
\bibliography{references}

\end{document}